\pdfoutput=1
\documentclass[11pt,letterpaper]{article}
\usepackage[margin=1in]{geometry}
\usepackage{times}
\usepackage{xcolor}
\usepackage{graphicx}
\usepackage{amsmath}
\usepackage{amssymb}
\usepackage{booktabs}
\usepackage{multirow}
\usepackage[numbers,sort&compress]{natbib}
\usepackage[font=small,labelfont=bf]{caption}
\usepackage[hyphens]{url}
\usepackage{microtype}
\graphicspath{{figures/}}
\definecolor{linkblue}{rgb}{0.21,0.49,0.74}
\usepackage[breaklinks,colorlinks,allcolors=linkblue]{hyperref}
\hypersetup{pdftitle={Parameter-Efficient Multi-Task Reinforcement Learning for Vision-Language-Action Models},pdfauthor={Xueyang Guo, Yihan Zhou, Jiayu Qu, Hongrui Zhu, Mingcong Li, Xu Yang, Jiale Chen, Yilin Mo}}

\title{Parameter-Efficient Multi-Task Reinforcement Learning\\for Vision-Language-Action Models}

\author{%
Xueyang Guo\textsuperscript{1,2}, Yihan Zhou\textsuperscript{1,2}, Jiayu Qu\textsuperscript{2,3}, Hongrui Zhu\textsuperscript{4},\\
Mingcong Li\textsuperscript{3}, Xu Yang\textsuperscript{1}, Jiale Chen\textsuperscript{5}, Yilin Mo\textsuperscript{1,2}\\[4pt]
{\normalsize\textsuperscript{1}Tsinghua University\quad
\textsuperscript{2}Dexteleop\quad
\textsuperscript{3}Beijing Institute of Technology}\\
{\normalsize\textsuperscript{4}Zhejiang University\quad
\textsuperscript{5}The Chinese University of Hong Kong}
}
\date{}

\begin{document}
\maketitle

\begin{abstract}
Reinforcement learning (RL) post-training improves vision-language-action (VLA) models from environment feedback, yet updating the entire multimodal model is costly. Freezing the backbone and training compact adapters reduces this overhead, but how adapter design governs VLA post-training remains poorly understood. We study parameter-efficient multi-task RL post-training for VLA models under matched experimental conditions. Apparent performance gaps across adapter parameterizations stem primarily from the per-update policy step size they induce rather than from adapter capacity: a fixed learning rate produces markedly different policy steps across structures, and this mapping drifts upward during training. Mechanistic analysis shows that sampling noise dominates single-step gradients, so parameterization mainly sets how a learning rate maps to a policy step, whereas across steps only a small shared signal accumulates, enabling a compact adapter aligned with the principal input directions of the pretrained weights to reach comparable success. Building on this diagnosis, we present PE-RL, which pairs online closed-loop step-size regulation toward a plateau target with zero-residual principal-direction initialization of a single low-rank adapter. Across distinct VLA architectures, RL algorithms, and multi-task manipulation benchmarks, PE-RL matches full-parameter RL while updating less than $0.5\%$ of the parameters.
\end{abstract}

\section{Introduction}
\label{sec:intro}

\begin{figure}[t]
\centering
\includegraphics[width=\textwidth]{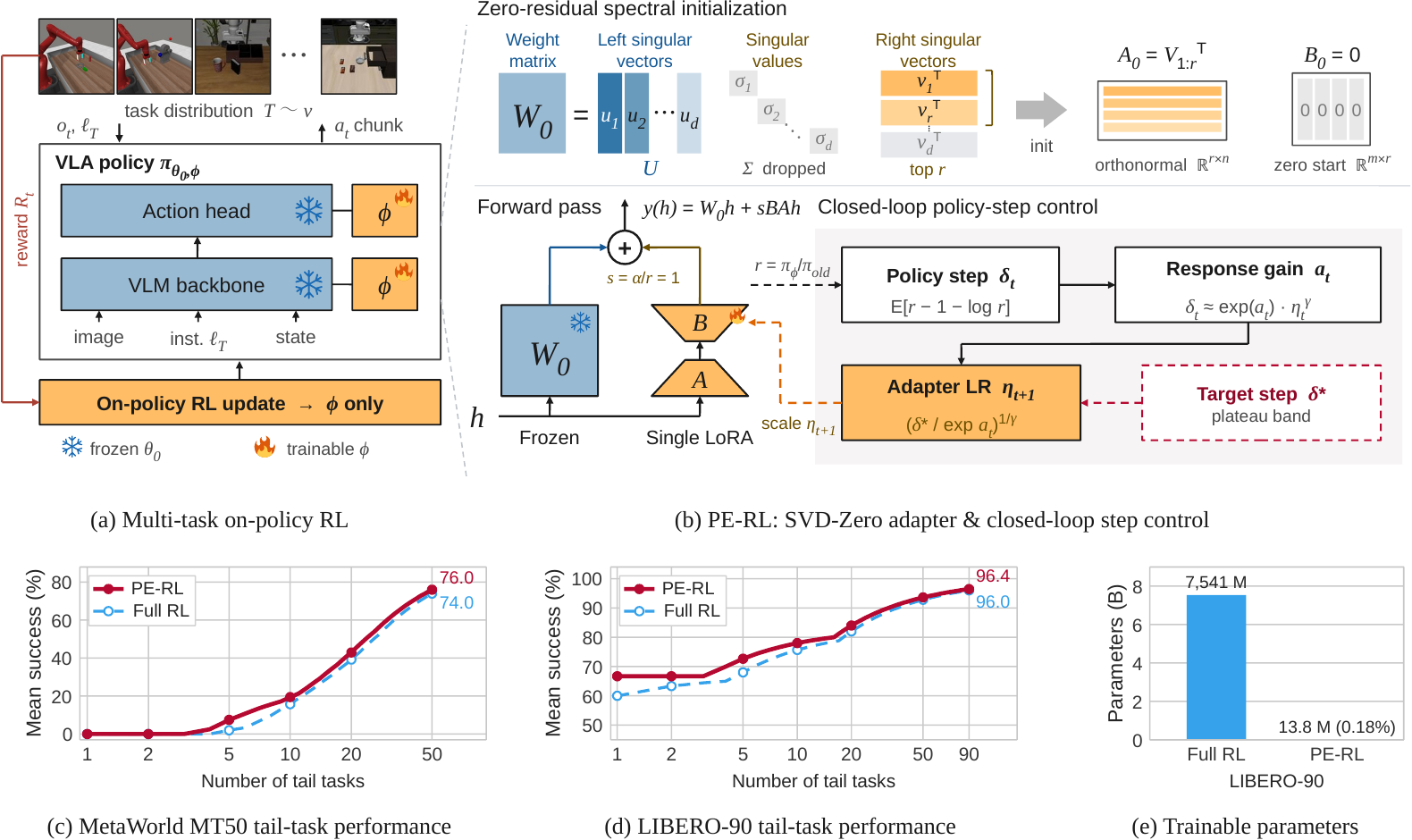}
\caption{\textbf{Overview of PE-RL.}
\textbf{(a)} Joint multi-task on-policy RL freezes the SFT backbone $\theta_0$ and updates only compact adapter parameters $\phi$ across tasks $T\sim\nu$.
\textbf{(b)} At each adapted projection $W_0$, SVD-Zero sets $A_0=V_{1:r}^\top$ from the leading right singular vectors of $W_0$ and $B_0=0$ to preserve the initial SFT policy, while a closed-loop controller tracks step sensitivity $a_t$ online to hold the realized KL step $\delta_t$ at a plateau target $\delta^*$.
\textbf{(c)}, \textbf{(d)} Mean success over the $n$ weakest tasks up to the full suite on MetaWorld MT50 and LIBERO-90.
\textbf{(e)} Trainable parameter count on LIBERO-90.}
\label{fig:main}
\end{figure}

Vision-language-action (VLA) models unify visual observation, language instruction, and action generation within a single generalist policy, transferring pretrained vision-language representations to robot control~\citep{zitkovich2023rt2,kim2025openvla,black2024pi0,physicalintelligence2025pi05}. Supervised fine-tuning (SFT) provides initial competence across manipulation tasks, yet policy performance remains bounded by demonstration coverage. Online reinforcement learning (RL) post-training improves pretrained policies further from their own rollouts~\citep{irevla2025,pirlflow2026,rlinfvla2025,pirl2025,qwenvla2026}, but updating the full multimodal model in every iteration incurs substantial gradient, optimizer-state, and checkpoint overhead. Parameter-efficient fine-tuning freezes the backbone and updates compact residuals~\citep{hu2022lora}, raising a central question: when post-training a VLA policy across dozens of manipulation tasks simultaneously, does restricting updates to a tiny adapter limit the performance reachable by multi-task RL?

Low-rank adaptation, spectral initialization, and conditional multi-expert routing have been studied mainly under supervised fine-tuning~\citep{meng2024pissa,dai2024deepseekmoe,ld_mole2026,vlague}. In RL, recent work shows that low-rank adapters can match full fine-tuning on language reasoning under per-configuration learning-rate sweeps~\citep{schulman2025lora} and benefit from geometry-preserving orthonormal initialization~\citep{rlpo2026,geora2026}, while in robotics low-rank RL has targeted sequential task acquisition~\citep{simplerecipe2026} or separate single-task policy libraries~\citep{memorypolicies2025}. Starting from a pretrained VLA policy, however, how adapter rank, initialization, multi-expert composition, and routing govern joint multi-task RL post-training, and why gaps arise under shared schedules or relative to full-parameter RL, remain open.

We compare adapter parameterizations against full-parameter updates in multi-task RL post-training of VLA policies under matched conditions, as summarized in Figure~\ref{fig:main}. Performance differences across adapter structures come primarily from the per-update policy step size they induce rather than from structural capacity: a shared learning rate yields markedly different policy steps across parameterizations, and the learning-rate-to-step sensitivity drifts upward throughout training. Probing the optimization dynamics reveals a noise-dominated, signal-limited regime. Within each iteration, sampling noise dominates policy gradients and coordinate-wise optimizer normalization equalizes step magnitudes across coordinates, turning parameterization into an exchange rate between learning rate and policy step size while penalizing oversized steps via diffusive drift. Across iterations, zero-mean sampling noise cancels while a small task-shared signal accumulates; once mixed-precision rounding truncation in full-parameter RL is removed, behavioral gains likewise reside in the linear projections, and a compact adapter aligned with their principal input directions co-adapts across layers to comparable success.

Guided by this diagnosis, we consolidate a minimalist recipe, PE-RL, which regulates the learning rate online to hold the per-update KL step within a stable plateau band and pairs it with principal-orthonormal zero-residual initialization~\citep{rlpo2026} of a single compact adapter, as shown in Figure~\ref{fig:main}. Selecting one target step from SFT per benchmark replaces separate searches over adapter structures, ranks, or learning rates, allowing PE-RL to match full-parameter RL across distinct VLA architectures, RL algorithms, and multi-task manipulation benchmarks while updating a tiny fraction of the parameters. Our contributions are threefold:
\begin{itemize}
\item We show that in joint multi-task RL post-training of VLA policies, apparent performance gaps across adapter ranks, multi-expert routers, and full-parameter updates stem primarily from the per-update KL shift of the policy rather than adapter capacity: a shared learning rate induces disparate, upward-drifting KL steps across structures, and all parameterizations collapse onto a shared performance plateau once KL steps are matched.
\item We explain why a compact principal-direction adapter on linear projections matches full-parameter RL: within each update, optimizer normalization of noise-dominated gradients equalizes per-parameter steps so adapter structure mainly rescales how learning rate maps to policy shift, whereas across updates noise cancels and task-shared signal accumulates. Moreover, once mixed-precision rounding no longer truncates small weight updates, full-parameter RL likewise concentrates its behavioral gains in the linear projections rather than modulation layers.
\item We distill this diagnosis into PE-RL, a minimalist recipe pairing online closed-loop KL step regulation with principal-orthonormal zero-residual initialization of a single low-rank adapter to match full-parameter RL across distinct policy architectures, RL algorithms, and multi-task benchmarks while updating under $0.5\%$ of the parameters and transferring a single target step across parameterizations.
\end{itemize}

\section{Related Work}
\label{sec:related}

\paragraph{Vision-language-action models and RL post-training.}
Pretrained VLA models such as RT-2, OpenVLA, OpenVLA-OFT, $\pi_0$, and $\pi_{0.5}$ transfer vision-language representations to robot control through SFT~\citep{zitkovich2023rt2,kim2025openvla,kim2025fine,black2024pi0,physicalintelligence2025pi05}, with multi-expert supervised adaptation explored in VLA-GSE~\citep{vlague}. Online RL post-training frameworks including iRe-VLA, $\pi_{\mathrm{RL}}$, RLinf-VLA, $\pi^{*}_{0.6}$, and Qwen-VLA improve manipulation policies further through environment interaction~\citep{irevla2025,pirlflow2026,rlinfvla2025,pirl2025,qwenvla2026}, typically updating full backbones or action experts, whereas prior low-rank robot RL studies focus on continual task sequences~\citep{simplerecipe2026} or per-task policy libraries~\citep{memorypolicies2025}. We examine joint multi-task RL post-training of a single generalist VLA policy and show that adapter parameterizations govern performance primarily through the per-update policy step size.

\paragraph{Low-rank parameterization, initialization, and expert routing.}
Supervised adaptation has motivated low-rank residuals~\citep{hu2022lora}, principal-component initialization in PiSSA~\citep{meng2024pissa}, rank-stabilized scaling in rsLoRA~\citep{kalajdzievski2023rslora}, and shared-plus-routed expert architectures~\citep{dai2024deepseekmoe,ld_mole2026,dikkala2023routing,moe_deep_rl}. Under RL, \citet{schulman2025lora} attribute low-rank sufficiency under tuned static learning rates to sparse $\mathcal{O}(1)$-bit episodic feedback, RLPO and GeoRA preserve initial policy geometry via orthonormal low-rank initialization for language-model reasoning~\citep{rlpo2026,geora2026}, and \citet{oh2026lowrank} audit singular spectra of full-parameter VLA RL checkpoints post hoc. Our analysis shows that in multi-task VLA RL, apparent multi-expert routing gains at fixed learning rates come from implicit step-size damping, upward curvature drift breaks static rate calibration, and mixed-precision storage rounding dictates where full-parameter RL concentrates its updates.

\paragraph{Trust-region methods and step-size regulation.}
Natural policy gradients define parameterization-invariant steps in distribution space~\citep{amari1998natural,kakade2001natural,kakade2002approximately}, TRPO and PPO constrain per-update divergence via trust regions, clipped ratios, or KL penalties~\citep{schulman2015trpo,schulman2017ppo}, and robot RL frameworks such as RSL-RL~\citep{rudin2022learning} adapt the learning rate of a fixed Gaussian policy toward a target KL. In parameter-efficient multi-task RL, we show that the learning-rate-to-step mapping spans orders of magnitude across adapter structures and drifts upward over training; regulating the realized KL step online aligns disparate parameterizations on a shared plateau without per-structure hyperparameter tuning.

\section{Step-Size Diagnosis in Multi-Task RL}
\label{sec:diagnosis}

Before introducing our method, we formalize multi-task RL post-training of VLA policies and examine why adapter ranks and multi-expert routers behave so differently under standard fixed learning-rate schedules.

\subsection{Problem Setup and Per-Update Policy Step}
\label{sec:method_formulation}
\label{sec:method_stepsize}
Consider a task set $\mathcal{T}$ with sampling distribution $\nu$, where each task $T\in\mathcal{T}$ provides a language instruction $\ell_T$. At decision step $t$, a VLA policy $\pi_{\theta_0,\phi}(\mathbf{a}_t\mid o_t,\ell_T)$ maps visual observation $o_t$ and instruction $\ell_T$ to an action chunk $\mathbf{a}_t$ and receives task reward. Starting from SFT weights $\theta_0$, parameter-efficient multi-task on-policy RL freezes $\theta_0$ and updates only adapter parameters $\phi$ inserted into the set $\mathcal{P}$ of linear projections across attention query, key, value, and output layers and MLP gate, up, and down layers, maximizing expected return across $T\sim\nu$ via PPO or GRPO.

At iteration $t$, let $\pi_{\mathrm{old}}$ denote the behavior policy before the update and $\pi_{\phi}$ the updated policy, with probability ratio $r = \pi_{\phi}(\mathbf{a}\mid o) / \pi_{\mathrm{old}}(\mathbf{a}\mid o)$. We quantify the per-update policy step size $\delta_t$ using the standard low-variance estimator of $\mathrm{KL}(\pi_{\mathrm{old}}\,\|\,\pi_{\phi})$ adopted in PPO implementations~\citep{huang2022cleanrl,raffin2021sb3}:
\begin{equation}
\delta_t = \mathbb{E}_{(o,\mathbf{a})\sim\pi_{\mathrm{old}}}\!\left[\,r - 1 - \log r\,\right],
\label{eq:k3}
\end{equation}
averaged across all micro-batches of iteration $t$. For nearby policies, this non-negative estimator has $\mathcal{O}(\delta_t)$ per-sample standard deviation, well below the $\mathcal{O}(\sqrt{\delta_t})$ standard deviation of $-\log r$. Unlike PPO clip fraction, which depends on the clipping threshold $\epsilon$, advantage sign, and epoch count and vanishes when updates stay well inside the trust region (for example single-epoch GRPO), $\delta_t$ measures action-distribution shift consistently across parameterizations and RL algorithms. In the local update regime, $\delta_t$ follows a power-law response to the policy learning rate $\eta_t$:
\begin{equation}
\delta_t \approx \exp(a_t)\,\eta_t^{\gamma},
\label{eq:powerlaw}
\end{equation}
where $\gamma > 0$ captures local curvature order and $a_t$ is the log step sensitivity at unit learning rate.

\begin{figure}[t]
\centering
\includegraphics[width=\textwidth]{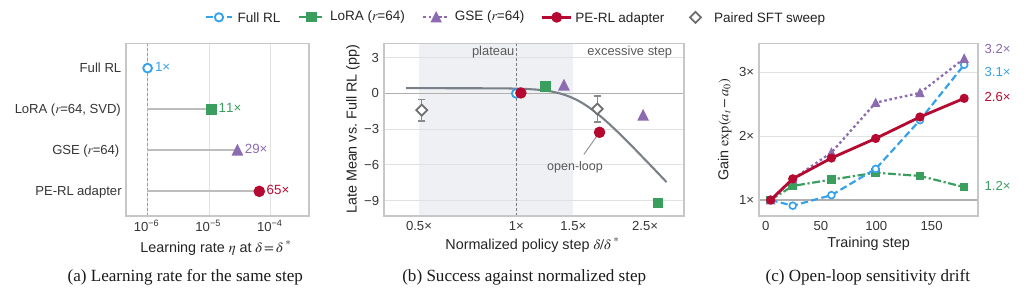}
\caption{\textbf{Shared step-size scale and open-loop sensitivity drift on MetaWorld MT50.}
\textbf{(a)} Applied learning rate $\eta$ needed to induce $\delta^*=0.0028$ under SVD-Zero adapters, spanning $65\times$ relative to Full RL.
\textbf{(b)} Late Mean difference relative to Full RL across normalized step size $\delta/\delta^*$, with open diamonds showing the 30-step paired SFT sweep relative to $1\times\delta^*$ and $95\%$ confidence intervals: all four parameterizations collapse onto one curve with a flat plateau over $0.5\text{--}1.5\times\delta^*$.
\textbf{(c)} Under open-loop cosine schedules, unit-rate sensitivity gain $\exp(a_t-a_0)$ drifts upward by $1.2\text{--}3.2\times$, pushing the one-shot calibrated PE-RL adapter past the plateau.}
\label{fig:collapse}
\end{figure}

\subsection{Policy Step Size Is the Common Scale}
\label{sec:analysis_collapse}
Varying adapter rank, initialization, or expert routing ostensibly alters model capacity, yet the quantity that directly governs each RL iteration is how far the update moves the action distribution, measured by $\delta_t$. To separate capacity from update scale on MetaWorld MT50, we compare full-parameter updates, denoted \emph{Full RL}, a single rank-64 \emph{LoRA} adapter, a rank-64 routed $1+7$ multi-expert \emph{GSE} adapter~\citep{vlague}, and a single rank-4 principal-direction adapter, denoted \emph{PE-RL adapter} and detailed in Section~\ref{sec:method}. Under a shared learning rate, these parameterizations induce markedly different policy steps; once compared at equal realized step size, inter-parameterization gaps disappear.

Figures~\ref{fig:collapse}a--b illustrate this collapse. Inducing the same moderate step $\delta^* = 0.0028$ under shared SVD-Zero initialization requires applied learning rates spanning $65\times$ in Figure~\ref{fig:collapse}a: relative to Full RL, rank-64 SVD-Zero LoRA requires $11\times$, rank-64 GSE requires $29\times$, and rank-4 PE-RL requires $65\times$. Conversely, at the same nominal learning rate, single LoRA takes a step $2.5\times$ larger than GSE because averaging one shared branch with renormalized top-2 routed branches implicitly damps the update by roughly $3\times$, explaining GSE's advantage over open-loop LoRA. Plotted against normalized step size $\delta/\delta^*$ in Figure~\ref{fig:collapse}b, Full RL, single LoRA, GSE, and the rank-4 adapter collapse onto a single response curve: Late Mean success and the paired SFT sweep remain flat across $0.5\text{--}1.5\times\delta^*$ and degrade broadly above $1.5\times\delta^*$. Across all 50 tasks, per-task success vectors among step-matched parameterizations correlate at $\rho = 0.93\text{--}0.98$, with residual differences matching the split-half noise floor of repeated runs within one parameterization.

\subsection{Step Sensitivity Drifts Under Open-Loop Schedules}
\label{sec:analysis_drift}
Because $a_t$ in Equation~\ref{eq:powerlaw} varies across parameterizations, a fixed learning rate confounds structural comparisons; while static scaling rules such as $s\propto 1/\sqrt{r}$~\citep{kalajdzievski2023rslora} or $1/r$~\citep{hu2022lora,schulman2025lora} adjust initial cross-rank scale, they address neither the adapter-to-Full-RL offset nor router branch damping. Even calibrating the initial learning rate once per structure is unreliable, because $a_t$ drifts upward throughout post-training. Under open-loop schedules in Figure~\ref{fig:collapse}c, unit-rate sensitivity gain $\exp(a_t - a_0)$ climbs to $1.2\times$ for rank-64 SVD-Zero LoRA, $2.6\times$ for the one-shot calibrated PE-RL adapter, $3.1\times$ for Full RL, and $3.2\times$ for GSE on MetaWorld. A pre-specified cosine decay happens to offset the mild $1.2\times$ drift of rank-64 LoRA, keeping it on the plateau, but under-compensates the $2.6\times$ drift of the rank-4 adapter and pushes it into the excessive-step regime of Figure~\ref{fig:collapse}b.

This upward drift persists well after optimizer moments equilibrate. Weight-perturbation probes show that the parameter update norm per unit learning rate stays constant from step~20 to step~200; instead, local KL curvature of the policy along active AdamW update directions is $20\text{--}240\times$ higher than along norm-matched random directions and grows by another $2\text{--}7\times$ as the policy evolves.

\section{PE-RL: Step-Regulated Principal-Direction Adaptation}
\label{sec:method}

The diagnosis in Section~\ref{sec:diagnosis} implies that no complex adapter architecture is needed once policy step size and read-subspace conditioning are controlled: PE-RL pairs online KL step regulation against sensitivity drift with policy-preserving principal-orthonormal initialization~\citep{rlpo2026} on a single compact adapter.

\subsection{Closed-Loop Step-Size Regulation}
\label{sec:method_controller}
To decouple update magnitude from parameterization scale and curvature drift, we estimate $a_t$ online and regulate $\eta_t$ toward a constant target $\delta^*$ inside the plateau band. After initializing $a_t$ as the sample mean over the first $K$ iterations, each iteration $t\ge K$ updates $a_t$ via an exponential moving average with weight $\lambda$:
\begin{equation}
a_t \leftarrow (1-\lambda)a_{t-1} + \lambda(\log\delta_t - \gamma\log\eta_t),
\label{eq:ema_intercept}
\end{equation}
and inverts Equation~\ref{eq:powerlaw} to set the learning rate for iteration $t+1$:
\begin{equation}
\eta_{t+1} = \left(\frac{\delta^*}{\exp(a_t)}\right)^{1/\gamma},
\label{eq:lr_inversion}
\end{equation}
with bounded per-step multiplicative changes for smooth transitions. At the fixed point of Equations~\ref{eq:ema_intercept}--\ref{eq:lr_inversion}, $\delta_t = \delta^*$ regardless of $\gamma$, which affects only the convergence rate following a transient deviation. Because multi-task RL for VLA policies exhibits a wide plateau over $0.5\text{--}1.5\times\delta^*$, $\delta^*$ only needs to sit inside this band: a short 30-step paired sweep from SFT on the adapter selects $\delta^*$ via the one-standard-error (1-SE) rule once per benchmark, and the same $\delta^*$ transfers directly across ranks, initializations, and full-parameter updates without retuning learning rates.

\subsection{Policy-Preserving Principal-Direction Adapter}
\label{sec:method_adapter}
With $\delta_t$ stabilized online, adaptation requires neither multi-expert compositions nor learned routers. Each frozen linear layer $W_0\in\mathbb{R}^{m\times n}$ in $\mathcal{P}$ receives a single low-rank residual with scale $s=\alpha/r$ and $A\in\mathbb{R}^{r\times n}$, $B\in\mathbb{R}^{m\times r}$:
\begin{equation}
y(h) = W_0 h + b_0 + s\,B A h.
\label{eq:single_lora}
\end{equation}
Absorbing singular values into nonzero factors, as in supervised spectral initialization~\citep{meng2024pissa}, amplifies early RL updates and degrades stability~\citep{rlpo2026}. Following principal orthonormal initialization~\citep{rlpo2026}---denoted \emph{SVD-Zero} below to contrast with random row-orthonormal \emph{Ortho-Zero} and \emph{Kaiming-Zero}---given $W_0 = U\Sigma V^\top$ ordered by descending singular values, we discard $\Sigma$, set the row-orthonormal read matrix $A_0$ to the leading $r$ right singular vectors, and initialize the write matrix $B_0$ to zero:
\begin{equation}
A_0 = \left[\,v_1,\, \ldots,\, v_r\,\right]^\top \in \mathbb{R}^{r\times n},
\qquad B_0 = 0 \in \mathbb{R}^{m\times r}.
\label{eq:svdzero}
\end{equation}
Initializing $B_0 = 0$ guarantees $\pi_{\theta_0,\phi_0} = \pi_{\theta_0}$, so rollouts start from the exact SFT policy. Under closed-loop step control, comparing SVD-Zero against Ortho-Zero and Kaiming-Zero separates row-orthonormal anchoring from principal-direction gradient capture, as analyzed in Section~\ref{sec:analysis_mechanism}. Both $A$ and $B$ update freely during training.

\section{Experiments}
\label{sec:experiments}

\subsection{Experimental Setup}
\label{sec:setup}
We evaluate joint multi-task RL post-training across two distinct VLA paradigms. On MetaWorld MT50~\citep{yu2020metaworld}, we train the flow-matching policy $\pi_{0.5}$ across all 50 tasks using Flow-SDE PPO, adapting all 126 linear projections of the action expert while keeping the vision-language prefix frozen; because evaluation accuracy plateaus by iteration~120 and continuation runs through 300 iterations remain flat within run-to-run noise, controlled comparisons standardize on a 200-iteration window. On LIBERO-90~\citep{libero}, we train the autoregressive policy OpenVLA-OFT across all 90 tasks for 300 iterations using critic-free GRPO, adapting the 437 linear projections across the full model in PE-RL and GSE. Every run starts from the shared SFT checkpoint of its benchmark and evaluates on 512 fixed-reset episodes spanning all tasks; we report peak checkpoint success over this window, denoted \emph{Peak}, which carries $\sim 3$~pp max-selection variance across checkpoints, alongside \emph{Late Mean}, the final five-checkpoint mean over steps 119--199 on MetaWorld and 220--300 on LIBERO-90 with a $\sim 1.5\text{--}2$~pp run-to-run band.

Tables~\ref{tab:metaworld_results}--\ref{tab:libero_results} compare five representative methods under matched rollout budgets, distilled from over 20 controlled multi-task RL configurations across adapter ranks, initializations, three multi-expert routing schemes, open- and closed-loop step schedules, and mixed- versus master-precision updates, alongside module-grafting probes and spectral audits across four manipulation suites in Sections~\ref{sec:ablation}--\ref{sec:analysis}. \textbf{SFT} is the initial supervised policy. \textbf{Full RL} updates all parameters in the adaptation scope; on MetaWorld, projections maintain FP32 master weights at matched step size to avoid the BF16 rounding truncation analyzed in Section~\ref{sec:analysis_precision}. \textbf{LoRA-RL ($r=64$)} trains a single rank-64 Kaiming-Zero adapter at the default open-loop learning rate. \textbf{GSE ($1+7, r=64$)}~\citep{vlague} combines one shared and seven routed SVD-Zero experts under top-2 hidden-state routing at total rank 64. \textbf{PE-RL} trains a single rank-4 SVD-Zero adapter under closed-loop KL tracking at $\delta^* = 0.0028$ per chunk on MetaWorld and $0.00128$ per token on LIBERO-90. In addition to overall accuracy and mean success on the ten weakest tasks, denoted \emph{Worst-10}, we report updated policy parameters, relative training FLOPs in the adapted modules, and analytic FP32 memory for gradients and AdamW states, denoted \emph{Mem.}

\subsection{Main Results: MetaWorld MT50}
\label{sec:metaworld}

\begin{table}[t]
\centering
\small
\caption{Joint post-training of $\pi_{0.5}$ across all 50 MetaWorld MT50 tasks. \emph{Peak} is the best evaluation checkpoint over the 200-iteration window, and \emph{Late Mean} averages the final five checkpoints over steps 119--199. \emph{Trainable} counts updated policy parameters, and \emph{Mem.} reports analytic FP32 gradient plus AdamW-state memory. \textsuperscript{*}Default open-loop learning rate $5\times 10^{-5}$.}
\label{tab:metaworld_results}
\setlength{\tabcolsep}{0pt}
\begin{tabular*}{\columnwidth}{@{\extracolsep{\fill}}lccccccc@{}}
\toprule
Method & Peak & Late Mean & W-10 & Trainable & Ratio & FLOPs & Mem. \\
 & (\%) & (\%) & (\%) & (M) & (\%) & (rel.) & (GB) \\
\midrule
SFT & 50.0 & --- & 1.0 & 0 & 0 & --- & --- \\
Full RL & 74.0 & 70.5 & 15.7 & 430.10 & 100 & 1.00 & 5.16 \\
LoRA-RL ($r=64$)\textsuperscript{*} & 63.2\textsuperscript{*} & 61.3\textsuperscript{*} & 4.5\textsuperscript{*} & 27.72 & 6.45 & 0.77 & 0.333 \\
GSE ($1+7, r=64$) & 74.9 & 68.7 & \textbf{23.0} & 29.14 & 6.78 & 0.78 & 0.350 \\
\textbf{PE-RL (Ours)} & \textbf{76.0} & \textbf{70.6} & 19.5 & \textbf{1.73} & \textbf{0.40} & \textbf{0.71} & \textbf{0.021} \\
\bottomrule
\end{tabular*}
\end{table}

Table~\ref{tab:metaworld_results} highlights three findings on MetaWorld MT50.

\paragraph{A single rank-4 adapter matches full-parameter RL.}
PE-RL raises peak success from $50.0\%$ to $76.0\%$ and Worst-10 success to $19.5\%$, matching Full RL in both Peak at $74.0\%$ and Late Mean at $70.6\%$ versus $70.5\%$, where both methods reach the same asymptotic ceiling imposed by binary sparse rewards on near-zero SFT tail tasks. It achieves this while updating $0.40\%$ of the active parameters, cutting adapted-module training FLOPs by $29.2\%$, and reducing gradient and optimizer-state memory from $5.16$~GB to $0.021$~GB.

\paragraph{Standard LoRA-RL stalls from excessive step size, not low capacity.}
Rank-64 LoRA-RL has $16\times$ the trainable parameters of PE-RL yet stalls at $63.2\%$ Peak, $61.3\%$ Late Mean, and $4.5\%$ on Worst-10 because the nominal learning rate of $5\times 10^{-5}$ drives realized policy steps to $2.75\times\delta^*$, far past the plateau in Figure~\ref{fig:collapse}b. Holding the Kaiming-Zero initialization fixed and adding closed-loop step control at rank~$4$, without retuning the learning rate, recovers Peak to $71.7\%$, Late Mean to $66.1\%$, and Worst-10 to $18.1\%$ in Group~C of Table~\ref{tab:ablations}: step regulation thus closes $4.8$ of the $9.3$~pp Late-Mean gap to PE-RL, with the remaining $4.5$~pp attributable to initialization, while the concurrent rank reduction is separately controlled in Group~B, where Late Mean varies by under $0.8$~pp across a $64\times$ rank range at matched step.

\paragraph{Multi-expert routing adds no capacity once step size is aligned.}
GSE updates $17\times$ more parameters than PE-RL yet reaches lower Peak at $74.9\%$ versus $76.0\%$ and lower Late Mean at $68.7\%$ versus $70.6\%$. Consistent with Section~\ref{sec:analysis_collapse}, GSE outperforms open-loop single LoRA early in training because expert averaging implicitly damps the initial policy step, before learned router updates push realized steps to $2.47\times\delta^*$ and lower Late Mean; regulating step size directly makes multi-expert routing unnecessary.

\subsection{Cross-Architecture and Cross-Algorithm Transfer: LIBERO-90}
\label{sec:libero}

\begin{table}[t]
\centering
\small
\caption{Joint post-training of OpenVLA-OFT across all 90 LIBERO-90 tasks with critic-free GRPO. \emph{Peak} is the best checkpoint over the 300-iteration window and \emph{Late Mean} averages the final five over steps 220--300; other columns follow Table~\ref{tab:metaworld_results}.}
\label{tab:libero_results}
\setlength{\tabcolsep}{0pt}
\begin{tabular*}{\columnwidth}{@{\extracolsep{\fill}}lccccccc@{}}
\toprule
Method & Peak & Late Mean & W-10 & Trainable & Ratio & FLOPs & Mem. \\
 & (\%) & (\%) & (\%) & (M) & (\%) & (rel.) & (GB) \\
\midrule
SFT & 80.1 & --- & 33.0 & 0 & 0 & --- & --- \\
Full RL & 96.0 & 95.3 & 75.7 & 7541.24 & 100 & 1.00 & 90.5 \\
LoRA-RL ($r=64$)\textsuperscript{*} & 92.1\textsuperscript{*} & 91.2\textsuperscript{*} & 61.3\textsuperscript{*} & 221.66 & 2.94 & 0.78 & 2.66 \\
GSE ($1+7, r=64$) & 95.4 & 94.6 & 69.0 & 232.28 & 3.08 & 0.78 & 2.79 \\
\textbf{PE-RL (Ours)} & \textbf{96.4} & \textbf{95.8} & \textbf{78.0} & \textbf{13.84} & \textbf{0.18} & \textbf{0.75} & \textbf{0.166} \\
\bottomrule
\end{tabular*}
\end{table}

We transfer PE-RL unchanged to OpenVLA-OFT, which decodes discrete action tokens, under critic-free GRPO on LIBERO-90. A 30-step paired adapter sweep from SFT selects $\delta^*=0.00128$ per token without architecture or learning-rate tuning. In Table~\ref{tab:libero_results}, PE-RL reaches $96.4\%$ Peak, $95.8\%$ Late Mean, and $78.0\%$ Worst-10, matching Full RL at $96.0\%$ and $95.3\%$ and GSE at $95.4\%$ and $94.6\%$ while updating $0.18\%$ of the $7.54$B parameters and cutting gradient and optimizer-state memory from $90.5$~GB to $0.17$~GB. Closed-loop step regulation with a single principal-direction adapter thus transfers cleanly across continuous flow-matching and discrete token policies as well as actor-critic and critic-free RL.

\subsection{Ablation Studies}
\label{sec:ablation}

\begin{table}[t]
\centering
\footnotesize
\caption{MetaWorld MT50 ablations under the shared 200-step protocol. \emph{Peak} reports the best evaluation checkpoint over the 200-iteration window, and \emph{Late Mean} averages the final five checkpoints over steps 119--199.}
\label{tab:ablations}
\setlength{\tabcolsep}{4pt}
\begin{tabular}{@{}llcc@{}}
\toprule
Ablation dimension & Configuration & Peak (\%) & Late Mean (\%) \\
\midrule
\multirow{5}{*}{A. Step-size control}
 & One-shot clip calibration over 10 steps, then open-loop cosine & 70.5 & 67.3 \\
 & One-shot KL calibration to $\delta^*$ over 10 steps, then open-loop cosine & 69.7 & 67.1 \\
 & Closed-loop clip-fraction tracking & 72.1 & \textbf{70.8} \\
 & RSL-RL adaptive learning rate~\citep{rudin2022learning} at KL target $\delta^*$ & 72.4 & 69.4 \\
 & \textbf{Closed-loop KL tracking at constant $\delta^*$ (PE-RL)} & \textbf{76.0} & 70.6 \\
\midrule
\multirow{3}{*}{B. Adapter rank}
 & Rank $r=1$ ($0.43$M parameters) & 72.7 & 70.3 \\
 & \textbf{Rank $r=4$ ($1.73$M parameters, PE-RL)} & \textbf{76.0} & 70.6 \\
 & Rank $r=64$, SVD-Zero, 3-probe tuned open loop ($27.72$M parameters) & 74.3 & \textbf{71.1} \\
\midrule
\multirow{3}{*}{C. Subspace initialization}
 & \textbf{SVD-Zero: leading right singular vectors of $W_0$ (PE-RL)} & \textbf{76.0} & \textbf{70.6} \\
 & Ortho-Zero: random row-orthonormal read subspace & 74.2 & 69.3 \\
 & Kaiming-Zero: random Kaiming read subspace & 71.7 & 66.1 \\
\midrule
\multirow{3}{*}{D. Target schedule}
 & \textbf{Constant $1.0\times\delta^*$: KL target $0.0028$ (PE-RL)} & \textbf{76.0} & \textbf{70.6} \\
 & Cosine-decaying KL target ($0.7\text{--}1.6\times\delta^*$) & 70.1 & 68.1 \\
 & Open-loop drift after 10-step calibration ($1.5\text{--}1.8\times\delta^*$) & 70.5 & 67.3 \\
\midrule
\multirow{2}{*}{E. Adapter structure}
 & Single rank-64 SVD-Zero LoRA, 3-probe step-matched & 74.3 & 71.1 \\
 & $1+7$ GSE ($r=64$) across frozen and learned hidden-state routers & 72.7--74.9 & 68.7--71.2 \\
\bottomrule
\end{tabular}
\end{table}

Table~\ref{tab:ablations} isolates each design dimension of PE-RL on MetaWorld MT50:
\begin{enumerate}
\item \textbf{Closed-loop regulation and step metric.} Fitting the learning rate over the first 10 iterations and running open loop thereafter lets realized steps drift past $1.5\times\delta^*$, dropping Late Mean to $67.3\%$ under clip calibration and $67.1\%$ under direct KL calibration to $\delta^*$, more than $3$~pp below closed-loop tracking at $70.6\%\text{--}70.8\%$. Confirming that realized step size rather than controller complexity governs convergence, the per-minibatch adaptive learning-rate rule of RSL-RL~\citep{rudin2022learning} at $\delta^*$ also recovers $69.4\%$ Late Mean against $67.1\%$ under open-loop drift, though multi-epoch updates induce an $11\times$ within-iteration learning-rate sawtooth with $51\%$ of iterations exceeding $1.5\times\delta^*$ versus $4\%$ under iteration-level tracking; clip tracking likewise reaches $70.8\%$ under PPO but cannot transfer to single-epoch GRPO where clip fractions vanish.
\item \textbf{Minimal rank sufficiency.} At matched policy steps, single adapters spanning a $64\times$ range in rank at $r=64$, $r=4$, and $r=1$ achieve Late Mean within $0.8$~pp of one another at $71.1\%$, $70.6\%$, and $70.3\%$.
\item \textbf{Orthonormal and principal-direction initialization.} At matched step size, row-orthonormal zero-residual initialization improves Late Mean substantially over Kaiming-Zero at $66.1\%$, reaching $69.3\%$ for Ortho-Zero and $70.6\%$ for SVD-Zero, a $+3.2$ to $+4.5$~pp gain exceeding run-to-run variation. Between the two orthonormal bases, SVD-Zero edges Ortho-Zero by $1.3$~pp in Late Mean within the single-run band, so we adopt deterministic $V_{1:r}^\top$ for its higher gradient capture without claiming a separate jump over random orthonormal directions.
\item \textbf{Target schedule and plateau width.} Holding a constant target at $1.0\times\delta^*$ reaches $70.6\%$ Late Mean in Group~D, whereas a cosine-decaying target that overshoots early to $1.6\times\delta^*$ drops to $68.1\%$ and open-loop drift drops to $67.3\%$. Across step-aligned 200-step runs, Late Mean stays flat over $1.0\text{--}1.5\times\delta^*$ at $69.4\%\text{--}71.2\%$, while the 30-step paired sweeps from SFT confirm flat bands down to $0.5\times\delta^*$ across $0.5\times$, $1\times$, and $2\times$ targets on MetaWorld at $53.7\%$, $55.1\%$, and $53.8\%$ and across $1\times$, $2\times$, and $4\times$ targets on LIBERO-90 at $81.0\%$, $83.2\%$, and $82.3\%$, with the 1-SE rule selecting $\delta^*$ on both benchmarks.
\item \textbf{Single adapter versus routed experts.} Aligning the step size of a single rank-64 LoRA with $1+7$ GSE brings the single adapter to $74.3\%$ Peak and $71.1\%$ Late Mean, matching frozen and learned hidden-state GSE variants at $72.7\%\text{--}74.9\%$ Peak and $68.7\%\text{--}71.2\%$ Late Mean and confirming that fixed-rate structural gaps are step-size confounds.
\end{enumerate}

\paragraph{Tuned standard LoRA versus PE-RL.}
Aligning step size alone on a rank-4 Kaiming-Zero LoRA reaches only $66.1\%$ Late Mean in Group~C, $4.5$~pp below PE-RL; reaching $71.1\%$ open loop at rank~64 in Group~B requires SVD-Zero initialization, three learning-rate probes, and a mild $1.2\times$ sensitivity drift that happens to match the pre-set cosine decay. At rank $r=4$ with $16\times$ fewer parameters, $2.6\times$ drift outpaces that same cosine schedule even after 10-step calibration and drops Late Mean to $67.1\%\text{--}67.3\%$ in Group~A, whereas closed-loop tracking holds $\delta^*$ across parameterizations without per-structure learning-rate search.

\section{Optimization Mechanism and Update Localization}
\label{sec:analysis}

Sections~\ref{sec:diagnosis}--\ref{sec:experiments} establish that matching per-update policy step size collapses structural gaps and allows a rank-4 adapter to match Full RL. We now examine the optimization mechanism behind single-step exchange rates and low-rank sufficiency, and resolve where full-parameter VLA RL writes its updates.

\subsection{Single-Step Noise Equalization and Cross-Step Accumulation}
\label{sec:analysis_mechanism}

\begin{figure}[t]
\centering
\includegraphics[width=\textwidth]{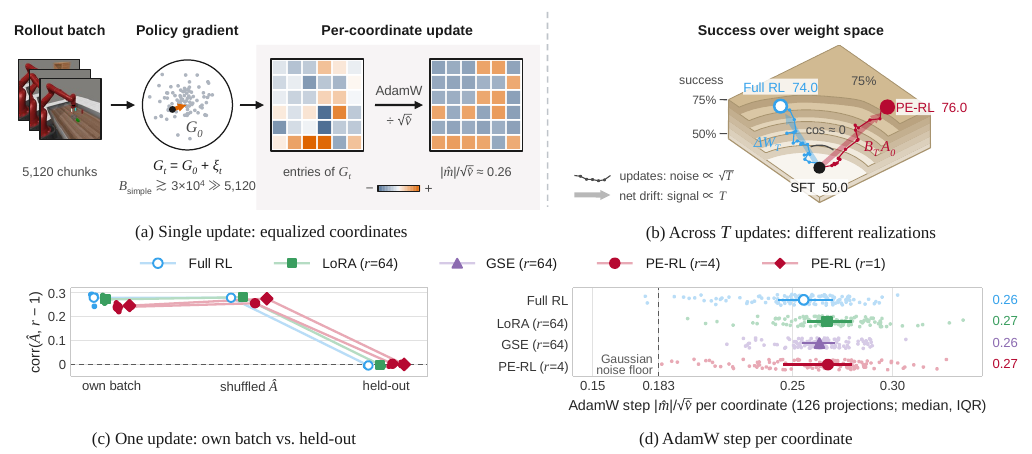}
\caption{\textbf{Single-step noise equalization and cross-step signal accumulation on MetaWorld MT50.}
\textbf{(a)}, \textbf{(d)} Within a single update, sampling noise dominates policy gradients, $\|\xi_t\|\gg\|G_0\|$, and AdamW normalizes median coordinate steps $|\hat m|/\sqrt{\hat v}$ across all 126 projections to $0.25\text{--}0.27$ near the Gaussian floor $0.183$, so parameterization only sets how many coordinates sum into $\delta_t$.
\textbf{(c)} Correlation between advantage $\hat A$ and policy shift $r-1$ after one PPO update at step 100: own-batch correlation matches shuffled advantages, whereas held-out correlation vanishes.
\textbf{(b)} Across $T$ updates, noise cancels as $\sqrt{T}$ while coherent shared signal accumulates as $T$, and PE-RL reaches a low-rank realization $B_T A_0$ distinct from Full RL's $\Delta W_T$ with comparable success.}
\label{fig:mechanism}
\end{figure}

Figure~\ref{fig:mechanism} separates single-iteration gradient statistics from cross-iteration signal accumulation to explain both the step-size exchange rate and the effectiveness of a compact principal-direction adapter.

\paragraph{Sampling noise equalizes single-step coordinate updates.}
Sparse episodic rewards carry only $\mathcal{O}(1)$ bits per episode~\citep{schulman2025lora}, and each single-iteration policy gradient $G_t = G_0 + \xi_t$ is dominated by sampling noise: on MetaWorld, the split-half gradient noise scale $B_{\mathrm{simple}}\gtrsim 3\times 10^4$ chunks~\citep{mccandlish2018empirical} exceeds the $5{,}120$-chunk rollout batch by nearly an order of magnitude, as sketched in Figure~\ref{fig:mechanism}a. The step-100 held-out probe in Figure~\ref{fig:mechanism}c shows the behavioral consequence directly: a single PPO update reaches $0.24\text{--}0.28$ correlation between advantage $\hat A$ and probability-ratio shift $r-1$ on its own batch, identical to sham updates trained on randomly shuffled advantages, while correlation on held-out rollouts vanishes at $-0.004\text{--}0.003$. Starting from zero residual with an anchored read subspace $A\approx A_0$ whose row-space retention is $0.993$, the write-factor gradient is the orthogonal projection $G_t A_0^\top$~\citep{rlpo2026}. Because projected gradients remain noise-dominated, AdamW normalizes every trainable coordinate across Full RL, rank-64 LoRA, GSE, and PE-RL to a median $|\hat m|/\sqrt{\hat v}$ of $0.25\text{--}0.27$, just above the Gaussian white-noise floor $0.183$ in Figure~\ref{fig:mechanism}d. Per-coordinate signal-to-noise ratio is thus largely invariant to parameterization: under unit adapter scale $s=1$, changing rank sums $r$ equalized coordinates so initial update norms grow as $\sqrt{r}$~\citep{kalajdzievski2023rslora}, expert averaging damps them, and oversized steps accelerate diffusive random walks in policy space.

\paragraph{Coherent shared signal and cross-layer co-adaptation.}
Across $T$ updates, independent sampling noise cancels as $\sqrt{T}$ whereas coherent task-shared signal accumulates as $T$, as illustrated in Figure~\ref{fig:mechanism}b. In unconstrained FP32 Full RL on MetaWorld, this shared signal skews the leading singular values of $\Delta W_T = W_T - W_0$ to a median stable rank $\|\Delta W_T\|_F^2/\sigma_1^2$ of $3.4$, while residual diffusion leaves a broad high-rank tail with 95\%-energy rank $r_{95}$ of $43\text{--}138$ on MetaWorld and $2{,}364\text{--}3{,}311$ on LIBERO-90. A low-rank adapter therefore does not win by approximating $\Delta W_T$ layer by layer: on LIBERO-90, truncating Full RL projections to rank 64 post hoc lowers grafted success from $94.1\%$ to $87.0\%$ as approximation errors compound across depth. Trained end to end, low-rank adapters exhibit near-zero weight cosine with Full RL and reproduce little to none of its fixed-state action shift, yet reach comparable success by co-adapting across frozen layers.

\paragraph{Orthonormal conditioning and principal input directions.}
Read-subspace conditioning still governs adaptation efficiency at matched step size. Enforcing $A_0 A_0^\top = I_r$ in Ortho-Zero and SVD-Zero anchors the read subspace with $0.993$ row-space retention and accounts for $+3.2$ to $+4.5$~pp in Late Mean over Kaiming-Zero in Group~C of Table~\ref{tab:ablations}. Among orthonormal bases, $V_{1:4}^\top$ holds $9.1\times$ the Full RL update energy density of a random four-dimensional subspace and captures $13\times$ more per-step gradient energy, consistent with SVD-Zero performing on par with or slightly above Ortho-Zero at $70.6\%$ versus $69.3\%$ Late Mean while providing a deterministic initialization.

\subsection{Mixed-Precision Rounding Gating in Full-Parameter RL}
\label{sec:analysis_precision}
Recent spectral audits~\citep{oh2026lowrank} report that full-parameter RL in $\pi_{0.5}$ concentrates updates almost entirely in AdaRMS timestep-modulation layers while leaving attention and MLP projections nearly unchanged and high-rank. We find that this concentration is not an algorithmic preference of RL, but a rounding-gate artifact of default mixed-precision training, which updates projections in place in BF16 while keeping timestep modules in FP32. Because per-step RL coordinate updates fall below the BF16 rounding unit, $91\%\text{--}97\%$ of projection entries remain bitwise identical to SFT across default MetaWorld, ManiSkill, and CALVIN checkpoints, erasing over $97\%$ of intended projection update energy, while $75\%\text{--}77\%$ remain bitwise unchanged in all-BF16 OpenVLA-OFT on LIBERO-90. Maintaining FP32 master weights for the MetaWorld projections restores update density to $93\%$ and projection energy share to $71.3\%$, matching their $72.4\%$ parameter share with the median stable rank of $3.4$.

Module-grafting evaluations across 512 episodes confirm that functional attribution reverses once rounding truncation is removed. In default BF16 Full RL, which reaches $71.2\%$ against $50.0\%$ for SFT, grafting only timestep modules onto SFT recovers $71.1\%$ whereas projections alone reach $57.4\%$, reproducing~\citet{oh2026lowrank}. In step-matched FP32-master Full RL at $74.0\%$, however, grafting only attention and MLP projections reaches $67.9\%$, retaining $75\%$ of the RL gain, while timestep modules alone drop to $49.6\%$, matching SFT. Without mixed-precision rounding truncation, full-parameter RL carries its behavioral gains in the attention and MLP projections rather than the timestep modules, consistent with adapting only those linear projections in PE-RL.

\section{Conclusion}
\label{sec:conclusion}
We studied parameter-efficient multi-task RL post-training for VLA models under matched conditions. Performance gaps across adapter ranks, multi-expert compositions, and routers stem primarily from the per-update policy step sizes they induce rather than from capacity, as the learning-rate-to-step mapping varies markedly across parameterizations and drifts upward during training. Optimizer normalization of noisy single-step gradients makes parameterization a step-size exchange rate, while across steps coherent shared signal accumulates in the linear projections, where full-parameter RL likewise carries its gains once mixed-precision rounding truncation is removed. Pairing online KL step regulation with policy-preserving principal-direction initialization in PE-RL matches full-parameter RL across VLA architectures, RL algorithms, and multi-task benchmarks while updating under $0.5\%$ of the parameters. Two online-RL constraints remain: environment interaction is sample-intensive for physical robots, and rollouts still execute the full frozen policy. Extending this framework to sample-efficient learning and sim-to-real transfer is a promising future direction.

\end{document}